\documentclass[10pt,twocolumn,letterpaper]{article}

\usepackage[pagenumbers]{cvpr} 

\usepackage{graphicx}
\usepackage{xcolor}
\usepackage{amsmath}
\usepackage{amssymb}
\usepackage{booktabs}
\usepackage{multirow}
\usepackage[inline]{enumitem}
\usepackage{caption}
\usepackage{subcaption}
\newcommand{\ours}{I2MVFormer\xspace}
\newcommand{\ourssummary}{SVSummary\xspace}
\newcommand{\myparagraph}[1]{\vspace{2pt}\noindent{\bf #1}}
\DeclareMathOperator*{\argmax}{arg\,max}
\def\x{\textbf{x}}
\def\y{\textbf{y}}
\def\vi{\textbf{v}}

\usepackage[pagebackref,breaklinks,colorlinks]{hyperref}

\usepackage[capitalize]{cleveref}
\crefname{section}{Sec.}{Secs.}
\Crefname{section}{Section}{Sections}
\Crefname{table}{Table}{Tables}
\crefname{table}{Tab.}{Tabs.}

\def\cvprPaperID{0000} 
\def\confName{CVPR}
\def\confYear{2023}

\begin{document}

\title{
I2MVFormer: Large Language Model Generated Multi-View Document Supervision for Zero-Shot Image Classification
}

\author{\vspace{0em}
\setlength\tabcolsep{0.1em}
\resizebox{\linewidth}{!}
{\begin{tabular}{ccccccc} 
Muhammad Ferjad Naeem$^{*1}$, & Muhammad Gul Zain Ali Khan$^{*2,3}$, & Yongqin Xian$^5$, & 
Muhammad Zeshan Afzal$^{2,3}$, & \\ Didier Stricker$^{2,3}$, & Luc Van Gool$^1$, & Federico Tombari$^{4,5}$ \tabularnewline
\end{tabular}}
\\
\renewcommand{\arraystretch}{0.5}
\begin{tabular}{ccccc} 
    $^1$\normalsize{ETH Zürich,} &  $^2$\normalsize{TUKL} & $^3$\normalsize{DFKI} & $^4$\normalsize{TUM} & $^5$\normalsize{Google} 
\end{tabular}
}%
\maketitle
\footnotetext[1]{First and second author contributed equally.} 
\begin{abstract}
Recent works have shown that unstructured text (documents) from online sources can serve as useful auxiliary information for zero-shot image classification. 
However, these methods require access to a high-quality source like Wikipedia and are limited to a single source of information. 
Large Language Models (LLM) trained on web-scale text show impressive abilities to repurpose their learned knowledge for a multitude of tasks.
In this work, we provide a novel perspective on using an LLM to provide text supervision for a zero-shot image classification model. The LLM is provided with a few text descriptions from different annotators as examples. The LLM is conditioned on these examples to generate multiple text descriptions for each class~(referred to as views).
Our proposed model, \ours, learns multi-view semantic embeddings for zero-shot image classification with these class views. 
We show that each text view of a class provides complementary information allowing a model to learn a highly discriminative class embedding. Moreover, we show that \ours is better at consuming the multi-view text supervision from LLM compared to baseline models.
I2MVFormer establishes a new state-of-the-art on three public benchmark datasets for zero-shot image classification with unsupervised semantic embeddings.
\end{abstract}

\section{Introduction}
\label{sec:intro}
In Zero-Shot Learning (ZSL), we task an image classification model trained on a set of seen classes to generalize to a disjoint set of unseen classes using shared auxiliary information. While there has been great progress made in the field, most works treat the auxiliary information to be fixed to a set of human-labeled attributes~\cite{xian2018zero,25_SUNdataset,26_wah2011caltech,farhadi2009describing}. While powerful, these attributes are hard to annotate and expensive to scale~\cite{song2018selective,yu2013designing}. Unsupervised alternatives to attributes rely on pretrained word embeddings which provide limited information about a class.
Recent works~\cite{naeem2022i2dformer, wikiacl, wikina, wikiless} show that text documents from internet sources like Wikipedia can provide great auxiliary information for ZSL. Since these web documents describe a queried class in detail, they provide more information for the ZSL model compared to word embeddings. However, these methods only rely on a single source of text documents like Wikipedia, which might not sufficiently represent all classes a model is faced with. 
Multiple sources of text documents of a class can provide complementary information for the ZSL model. For example, in the case of birds, one source might focus more on the patterns of the feather, while another source might better describe the belly and the face of the bird. However, finding multiple good sources of text documents for each class requires additional annotation effort. 


\begin{figure}
    \centering
    \includegraphics[width= \columnwidth]{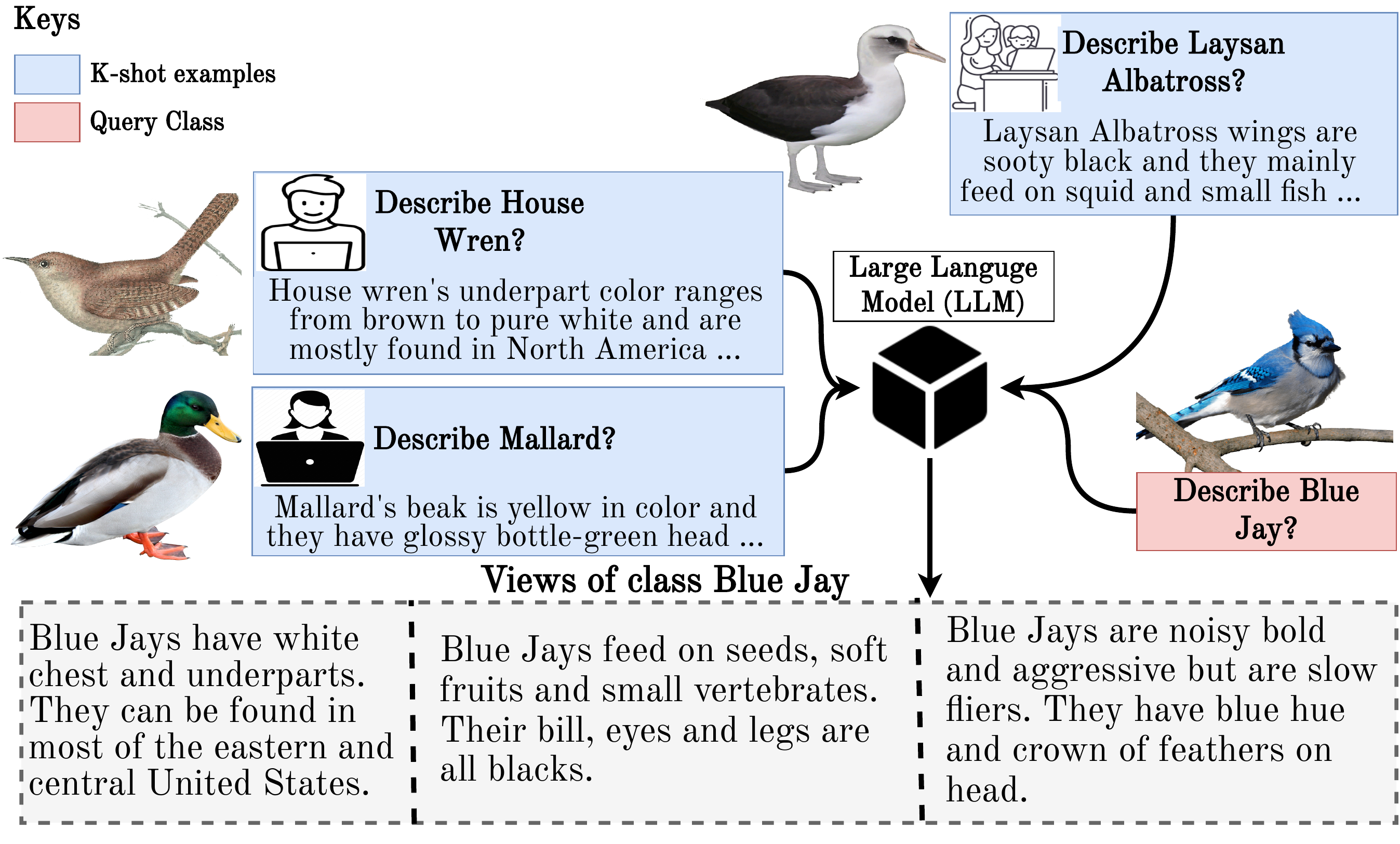}
    \caption{
    Different annotators focus on different attributes when describing a class.
    Large Language Models prompted with each of these annotations as k-shot examples can reveal complementary information about a class for zero-shot image classification. We refer to multiple LLM-generated descriptions as \emph{views} of a class.
    }
    \vspace{-5pt}
    \label{fig:teaser}
\end{figure}

Large Language Models (LLM)~\cite{gpt3, chowdhery2022palm, opt} trained on web-scale text have shown impressive abilities of using their learned information to solve a multitude of tasks. These models can be conditioned with a k-shot prompt to generalize to a wide set of applications~\cite{gpt3, mokady2021clipcap, zeng2022socratic} using knowledge from multiple sources they were trained on. In this work, we aim to generate multiple text descriptions of a class, that we recall as ``views" hereinafter, with an LLM using a k-shot prompting strategy. We show that the LLM can act as a mixture of annotators conditioned on different annotation styles to generate complementary information about a class. Moreover, we propose a novel model, \ours, which utilizes our memory-efficient summary modules to extract discriminative information from each view of a class with the aim of learning a multi-view class embedding.

Our contributions in this work are as follows.
\begin{enumerate*}[label={\arabic*)}]
    \item We provide the first study into using an LLM to generate auxiliary information for zero-shot image classification. Moreover, we propose a prompting strategy to extract multiple descriptions from an LLM that reveal complementary information about a class. 
    \item We propose \ours, a novel transformer-based model for zero-shot image classification which exploits multiple complementary sources of text supervision to learn a class embedding. 
    \ours utilizes our Single-View Summary~(SVSummary) module to extract rich discriminative information from each class view. This information is utilized by our Multi-View Summary~(MVSummary) module to represent a class-level set of tokens from multiple views. The multi-view tokens are aligned with the image to maximize global and local compatibility between the images and the multiple views.
    \item Our \ours achieves significant performance gains to establish a new state-of-the-art~(SOTA) in unsupervised class embeddings in ZSL on three public benchmarks AWA2~\cite{32_awa}, CUB~\cite{26_wah2011caltech} and FLO~\cite{OxfordFlowersDataset}.
\end{enumerate*}

\section{Related Work}
\label{sec:related_work}
\myparagraph{Zero-shot learning} aims to learn a model that can generalize beyond the seen classes it was trained on. This is accomplished by using side information that is shared with a set of disjoint unseen classes. Towards this, several methods learn a compatibility function between the image feature and a class embedding representing the auxiliary information~\cite{romera2015embarrassingly,CCGS16,DEM,xian2016latent,cape, compcos}. 
These methods often suffer from bias against unseen classes. To address this, another family of methods additionally learns the distribution of the features of images using a generative model~\cite{xian2018zero,brendel2019approximating,ABP,zslgan,kumar2018generalized,schonfeld2019generalized}. Approaches in this category focus on learning a class conditional generator that can generate features of unseen classes~\cite{kumar2018generalized,brendel2019approximating} or exploit semantic information about a class to generate features of unseen classes directly~\cite{schonfeld2019,zslgan}. Once learned on seen classes, these models generate features of unseen classes using its auxiliary information to tackle the bias issue. Other works focus on learning improved visual-semantic embeddings~\cite{liu2018generalized, DEM,jiang2019transferable, cacheux2019modeling} and training better image feature extractors~\cite{ji2018stacked,SGMA, apn}.
However, all these methods assume that the set of auxiliary information is fixed to human-labeled attributes~\cite{xian2018zero,25_SUNdataset,26_wah2011caltech,farhadi2009describing, c3d}. However, labeling attributes is expensive and hard to scale on large datasets as it requires expert annotators~\cite{song2018selective,yu2013designing,26_wah2011caltech}. 

\myparagraph{Unsupervised semantic embedding} aims to learn the
semantic embedding of seen and unseen classes using side information that does not require human intervention.
The most influential works in this direction use word embeddings from a pretrained model to encode semantic similarities~\cite{yamada2018wikipedia2vec,glove,socher2013zero,word2vec} and refine them using knowledge graphs~\cite{wang2018zero,kampffmeyer2019rethinking,bucher2017generating, cge, cocge}. VGSE\cite{vgse} learns a class embedding by using image patches and class embedding vectors. Several works have explored leveraging text documents from sources like Wikipedia to learn class embeddings since they contain rich information about a class. The literature in this direction exploits pretrained language models ~\cite{write , wikiatt, wikiless, wikiparts, zslgan, wikina, wikiacl} in addition to noise minimization through a predefined vocabulary\cite{wikiless} or part detection network~\cite{wikiparts, zslgan}. However, these works treat the embedding of the document with a pretrained model as fixed. Recently I2DFormer\cite{naeem2022i2dformer} propose a transformer-based model that learns a class embedding from raw text. Unlike zero-shot transfer models like CLIP\cite{clip}, which only maximizes the global compatibility between an image and text embedding, I2DFormer maximizes both the global and the local compatibility of the text features against the image. However, I2DFormer relies on expensive local attention between each image patch and document token, which does not scale well to large text sources.

\myparagraph{Large Language Models (LLM)} like GPT-3\cite{gpt3}, OPT\cite{opt} and PaLM\cite{chowdhery2022palm} are trained on very large web-scale datasets. Once trained, these models have impressive abilities towards zero-shot and few-shot inference on a multitude of tasks such as Open Question and Answering~\cite{qna}, generating code~\cite{codex}, text summarization~\cite{m2020holms}, etc. These models rely on a k-shot prompt defining the problem they have to solve along with 0 or k examples. Once prompted for the target class, they use their stored knowledge from web-scale training to generate text for the target class. Recently some works have tried to pair these models with vision models using generated text \cite{zeng2022socratic} or adding vision as a modality\cite{alayrac2022flamingo}. Other works explore prompting vision-language models~\cite{wang2022learning,li2022bridgeprompt,mokady2021clipcap} for continual learning, image caption generation, and action understanding. However, no work has yet explored leveraging an LLM to generate auxiliary information for zero-shot image classification.


\begin{figure*}
    \centering
    \includegraphics[width=\textwidth]{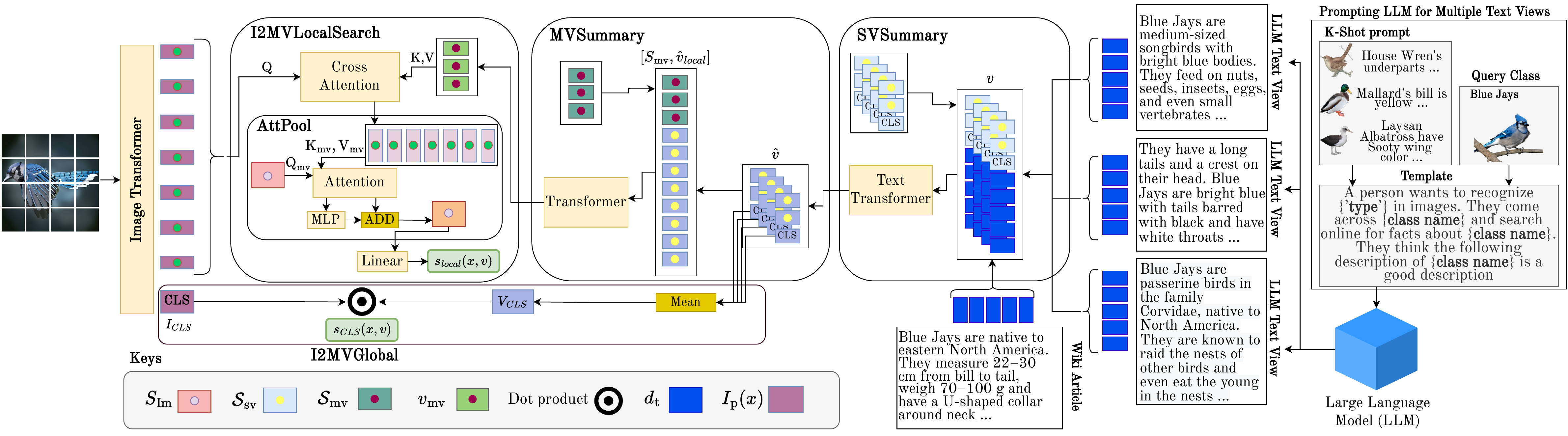}
    \caption{\textbf{\ours}, our novel Transformer based model uses LLM-generated document supervision conditioned on multiple annotators and noisy Wiki documents as multiple views of a class to learn a zero-shot model. Our SVSummary module generates a single view level $\mathtt{CLS}$ representation and local summary. The CLS summaries over multiple views are used to align the global features against the image. Our MVSummary module uses the summary tokens of each view to generate a multi-view summary of the class. This multi-view summary is aligned with the per-patch feature of the image in our I2MVLocalSearch module. Together the two modules learn a highly discriminative multi-view class embedding. 
    }
    \label{fig:model}
\end{figure*}

\section{Generating multiple text views with an LLM}
\label{subsec:prompt}

We define a text description of a class as its ``view". This is analogous to a document used in previous work\cite{naeem2022i2dformer}.
A single view from a source like Wikipedia as used in previous works\cite{naeem2022i2dformer, wikiacl, wikiatt, wikiless}, while powerful, can present knowledge gaps for less frequent classes. These classes might be described better in other, more specialized databases on the internet.
We propose to use LLMs to generate multiple views of a class, exploiting the property of these models of storing the knowledge from multiple internet sources.
In this section, we describe our novel prompting method to automatically generate multi-view text descriptions of object classes using a pretrained LLM.

While LLMs have impressive zero-shot abilities, practically, they require significant prompt engineering to get good zero-shot output, as noted in several works \cite{zhou2022cocoop,zhou2022coop,li2022bridgeprompt,wang2022learning,lee2021prompt_dialogue_history,Lee2021prompt_sentiment_analysis,Lu2022PromptDistributionlearning}. 
Since LLMs are few-shot learners\cite{gpt3}, given k-shot examples, they can generate output for any class they are prompted for.
We exploit this few-shot ability of LLMs to generate multiple text descriptions of each class representing multiple views. Given a set of annotated examples of how to describe a few classes of a dataset in natural language, the LLM can be prompted to mimic the labeling style to generate text supervision for all classes in a dataset. 
We require $f+1$ annotated examples to generate $f$ views from multiple inferences of the LLM. The extra example is reserved for replacement whenever the query class is present in the k-shot examples. We observed that without this, the LLM is susceptible to producing the exact class description as the one in the examples. 

To curate $f+1$ examples without requiring significant annotation cost, we rely on the Wiki articles released by \cite{naeem2022i2dformer}. These articles are filtered only to contain sections of Wikipedia labeled to contain visual information. This is done as sources like Wikipedia contain a lot of noise in the form of non-visual information. We notice that different wiki articles denote different annotation styles as the content is sourced from a pool of volunteers. In our work, we use $f=3$ as the number of views generated by LLM for each class. We randomly select four classes from each dataset and curate their wiki articles from \cite{naeem2022i2dformer} to only focus on the visually relevant details.
This allows us to obtain a relatively rich source of auxiliary information representing multiple sources without drastically increasing annotation effort. We now want the LLM to use these as examples of class descriptions to generate similar examples over all classes in the dataset. We append each labeled example with the following prompt. 

\textit{``A person wants to recognize \{'\textbf{type}'\} in images. They come across \{\textbf{class name}\} and search online for facts about \{\textbf{class name}\}. They think the following description of \{\textbf{class name}\} is a good description."}

In the given template, we use type as ``animals", ``birds" and ``flowers" for AWA2~\cite{xian2018awa2}, CUB~\cite{26_wah2011caltech} and FLO~\cite{OxfordFlowersDataset}, respectively, moreover \{\textbf{class name}\} defines the name of the labelled class. 
The target class is then entered into the template and fed as the input to the LLM with the k-shot examples. The LLM generates a description for this target class conditioned on the labeling style of our k-shot examples. We use $k=2$ for most of our experiments, i.e., each view is generated conditioned on 2 labeled examples. These 2 shots are a combination of the $3+1$ labeled examples for the 3 views of each class. We later show in supplementary that the LLM is fairly robust to the choice of prompt given the same k-shot example. 

\section{\ours}
\label{sec:method}
Most methods in zero-shot literature either rely on human-labeled attributes~\cite{xian2018zero,25_SUNdataset,26_wah2011caltech,farhadi2009describing, c3d} or pretrained word embeddings~\cite{yamada2018wikipedia2vec,socher2013zero, wang2018zero,kampffmeyer2019rethinking,bucher2017generating, cge, cocge} as auxiliary information. Recently I2DFormer~\cite{naeem2022i2dformer} has shown that text can be a powerful substitute without requiring significant labeling effort. 
Multiple descriptions, referred to as views in this work, of a class can provide complementary information of each class to the model and potentially lead to better performance. 
Our proposed model \textbf{Image to Multi-View Transformer~(\ours)} is designed to exploit the knowledge in multiple views of each class to learn a zero-shot model. \ours consists of two streams of transformers to process the visual and the textual data as shown in Figure~\ref{fig:model}. \ours aligns the global information available in each view with the image along with the fine-grained information. Moreover, \ours reduces the memory footprint of I2DFormer while significantly outperforming it on three public benchmarks.


\myparagraph{Notations.}
\label{subsec:notations}
We represent the classes in the training set as $\mathcal{Y}^s$ and the unseen classes only available at test time as $\mathcal{Y}^u$. 
Let $\mathcal{T}=\{(\x, \y, \vi) | \x \in \mathcal{X}^s, \y\in \mathcal{Y}^s, \vi\in \mathcal{V}^s \}$ be our training set where $\x$ denotes an RGB image from the training images $\mathcal{X}^s$, $\y$ is its label belonging to the seen classes $\mathcal{Y}^s$, $\vi=\{d_1, d_2 .... d_q\}$ is the set of documents representing $q$ views of the class and $\mathcal{V}_s$ the set of views of the seen classes. At test time an additional set $\mathcal{V}_u$ is made available as the views of novel classes where $\mathcal{V}=\mathcal{V}_s + \mathcal{V}_u$ and $\mathcal{V}_s \cap \mathcal{V}_u = \emptyset$. The ZSL task requires the model to predict a class from the set of unseen classes $\mathcal{Y}_u$, and the Generalized ZSL~(GZSL) requires the model to predict a class over both seen and unseen classes $\mathcal{Y}=\mathcal{Y}^s+\mathcal{Y}^u$.

\subsection{Image Transformer}
On the image side, our model learns $\mathcal{F}$, an image transformer as an embedding function. Given an image $\x \in \mathbb{R}^{H\times W \times C}$, we reshape it into a sequence of flattened 2D patches $\x_p \in \mathbb{R}^{N \times (P^2 C)}$, where $(H, W$) is the size of an input image with $C$ as the RGB channels, $(P, P)$ is the size of each image patch, and $N = HW/P^2$ is the resultant number of patches. Moreover, we append a $\mathtt{CLS}$ token to $\x_p$ as the input to the image transformer to learn a global image representation. Inspired by recent advances in multimodal learning~\cite{zhai2021lit}, we use a pretrained frozen image transformer~\cite{vit} followed by a learnable MLP layer which maps the features to a joint image-text embedding space with dimentionality $r$. 
$\mathcal{F}$ outputs $I_{CLS}(\x) \in \mathbb{R}^{r}$ as the global image feature and $I_\text{p}(\x) \in \mathbb{R}^{N\times r}$ as the patch-wise image embedding for the input image.

\subsection{\ourssummary: Extracting class level summary from each view. }
In our multi-view setup, each training class $\y$ is associated with $\vi$, the set of documents representing multiple views of the class. The model is now faced with an increased amount of text compared to existing works that rely on a single view. This makes existing solutions~\cite{naeem2022i2dformer} that rely on exhaustive attention on all document tokens against the image computationally very expensive. Our SingleViewSummary~(SVSummary) module is a text transformer that aims to represent the most discriminative information available in each text description (view) of a class into a fixed set of Summary tokens.
It is important to note that we define a summary as the most important features described in the text rather than its classical definition of a human consumable shortened description of the text.

Given $\vi=\{d_1, d_2,..., d_n\}$, the set of documents representing views of class $\y$, we pre-tokenize each document and represent it by a pretrained word embedding model similar to ~\cite{naeem2022i2dformer}. We learn a shallow MLP on top to improve upon the pretrained word embedding model. The output of this yields $d_{\text{t}} \in \mathbb{R}^{M \times r}$ for each document $d_{\text{t}}$ in the view set where $M$ is the length of the document and $r$ is the feature dimension. We introduce Single-View Summary tokens $\mathcal{S}_{\text{sv}} \in \mathbb{R}^{T \times r}$ as a set of $T$ learnable tokens. These tokens are introduced to specialize them for the downstream image compatibility task instead of using all $M$ tokens of a view. Given $T < M$, this results in a reduced constant memory cost of the later cross-modal alignment independent of the length of each view. $\mathcal{S}_{\text{sv}}$ is appended to each view and passed through a learnable text transformer with several Transformer encoding blocks. On the output side, we take the token representations corresponding to $\mathcal{S}_{\text{sv}}$ for every single view $d_{\text{t}}$ to get its summary $S_{\text{sv}}(d_{\text{t}})$. For $\vi$, the set of views of a class, this step yields $\hat{\vi} = \{\mathcal{S}_{\text{sv}}(d_1), \mathcal{S}_{\text{sv}}(d_2),...,\mathcal{S}_{\text{sv}}(d_q)\} \in \mathbb{R}^{q\times T \times r}$ as the learnable summary for each view of a class. The first summary token of each view is specialized as the $\mathtt{CLS}$ token, which contains the global feature of this view for global alignment in the next step. 

\subsection{I2MVGlobal: Learning global compatibility between image and multiple views}
I2MVGlobal aims to align the global feature of an image $\x$ with the ground truth class $\y$ using the view set $\vi$.
To accomplish this, we use the $\mathtt{CLS}$ token of each view in Figure~\ref{fig:model} and take a mean over the view dimension $q$ to get a global feature across all views defined as $V_{CLS}(\vi) \in \mathbb{R}^r$. We align this feature with the global image-level feature $I_{CLS}$ from the image encoding function $\mathcal{F}$. To accomplish this, we define a scoring function $s_{CLS}$ that measures the compatibility between $I_{CLS}$ and $V_{CLS}$ as a dot product:
\begin{equation}
 \label{eq:compatibility}
    s_{CLS}(\x, \vi) = I_{CLS}(\x) \cdot V_{CLS}(\vi).
\end{equation}
The learning objective aims to assign high values to correct pairs and low scores otherwise. For a particular training instance $(\x, \y, \vi)$, we minimize the following cross-entropy loss over $\mathcal{V}_s$ the set of views for the seen classes.
\begin{equation}
\label{eq:cls}
\begin{split}
    L_{CLS} = 
    -\log \left( \frac{\exp{s_{CLS}(\x, \vi)}}
    {\sum_{\vi'\in \mathcal{V}^{s}} \exp{s_{CLS}(\x, \vi')}}  \right)
\end{split}
\end{equation}




\subsection{MVSummary: Generating multi-view summary from all views}
Previous work in ZSL~\cite{naeem2022i2dformer} has shown that only aligning the global feature of an image to a text description can lead to over-fitting to seen classes. The existing solution to mitigate this relies on expensive cross-attention between all image and document tokens which becomes increasingly expensive as increased text is available for each class.
Our SVSummary module allows this at reduced memory cost as it generates a fixed number of summary tokens irrespective of the length of a view.

We concatenate the summary tokens of each view (except the $\mathtt{CLS}$ token, which is used in I2MVGlobal) in the view dimension $q$ to define $\hat{\vi}_{local} \in \mathbb{R}^{ q (T - 1) \times r}$, the summary of all views. This set will grow linearly, given increased views of each class, and can become expensive to align against the local features of an image. We mitigate this through our {Multi-View Summary Transformer (MVSummary)} module, which introduces a set of multi-view summary tokens $\mathcal{S}_{\text{mv}}\in \mathbb{R}^{T \times r}$ with $T$ learnable tokens. These tokens are aimed at summarizing the local summaries of all views into one class-level multi-view summary. We concatenate $\mathcal{S}_{\text{mv}}$ with $\hat{\vi}_{local}$ and use it as input to a learnable Transformer encoder block. The output feature representation corresponding to tokens in $\mathcal{S}_{\text{mv}}$ represent the multi-view local summary $\vi_{\text{mv}} \in \mathbb{R}^{T \times r}$ for a given class.

\subsection{I2MVLocalSearch: Fine-grained alignment between image and summary}
Our I2MVLocalSearch aims to align the patch-wise image features $I_\text{p}(\x) \in \mathbb{R}^{N\times r}$ with the multi-view summary tokens of each class $\vi_{\text{mv}}$. The core idea is that each summary token specializes in an aspect of a class defined by the multiple views of the class. Aligning an image to these encourage the model to focus on local features that are important for fine-grained classification. We define this as a query search problem where $Q= I_\text{p}(\x) W_q$ defines the visual query set, and $K = \vi_{\text{mv}}(\vi) W_k$, $V =\vi_{\text{mv}}(\vi) W_v$ define the keys to compare with and values to combine with. $W_q$, $W_k$ and $W_v$ are learnable linear mappings $\in \mathbb{R}^{r \times r}$. The first step involves computing an attention matrix between $Q$ and $K$ and using it to calculate a new multi-view representation of image patches $I_{\text{mvpatch}}= \mathrm{softmax}(\frac{QK^T}{\sqrt{r}})V\in \mathbb{R}^{N \times r}$.

We want to align this representation with the ground truth class $\y$. We define a learnable attention pooling to get an image-level feature from these patch representations.
We use a learnable Image summary token $\mathcal{S}_{\text{Im}} \in \mathbb{R}^{r}$ as the query $Q_{\text{mv}}$.
The keys $K_{\text{mv}}$ and values $V_{\text{mv}}$ are defined by passing $I_{\text{mvpatch}}$ through a linear layer. We compute attention between the query and key and use these values to compute intermediate representation $I'_{\text{mv}} \in \mathbb{R}^r$. This is passed through a learnable two-layer MLP and a skip connection to get the multi-view image feature $I_{\text{mv}} \in \mathbb{R}^r$ as:
\begin{equation}
\label{eq:att}
\begin{split}
    I'_{}\text{mv}(\x, \vi) &= \mathrm{softmax}(\frac{Q_{\text{mv}}K_{\text{mv}}^T}{\sqrt{r}})V_{\text{mv}}\\
    I_{\text{mv}}(\x, \vi) &= I'_{\text{mv}}(\x, \vi) + \mathrm{MLP}(I'_{\text{mv}}(\x, \vi))
\end{split}
\end{equation}
The multi-view image feature is used to predict a local alignment score $s_{local}$ by a learnable linear layer $J\in \mathbb{R}^{1\times r}$. This score is optimized with a cross-entropy loss $L_{local}$,
\begin{equation}
\label{eq:localloss}
\begin{split}
    s_{local}(\x,\vi) &= J(I_{\text{mv}})\\
    L_{Local} &= 
    -\log ( \frac{\exp{s_{local}(\x, \vi)}}
    {\sum_{\vi'\in \mathcal{V}^{s}} \exp{s_{local}(\x, \vi')}}  )
\end{split}
\end{equation}
To summarize, \ourssummary allows each document view a chance to describe the class. The MVSummary uses these to generate a multi-view class summary which is subsequently used by our I2MVLocalSearch. This multi-view summary is aligned with the patch-level image features to promote fine-grained feature learning. 

\subsection{Inference.}
Given an input image $\x$, a prediction $\hat{\y}$ corresponds to the view set that yields the highest compatibility score among unseen classes for ZSL and among both seen and unseen classes for GZSL: 
\begin{equation}
    \hat{\y} = \argmax_{\vi' \in \mathcal{V}} s_{CLS}(\x, \vi').
\end{equation}

\setlength{\tabcolsep}{4pt}
\renewcommand{\arraystretch}{1.2} 
\begin{table*}[t]
\centering
\small
\setlength{\aboverulesep}{0pt}\setlength{\belowrulesep}{0pt}
 \resizebox{\linewidth}{!}{%
   \begin{tabular}{ l l c c c c c c  c c c  c c c }
  	\toprule
  	& & \multicolumn{3}{c}{\textbf{Zero-Shot Learning}} & \multicolumn{9}{c}{\textbf{Generalized Zero-Shot Learning}} \\
  	\cmidrule(lr){3-5} \cmidrule(lr){6-14}
  \textbf{Model}	&  \textbf{Auxiliary Information} & \textbf{AWA2} & \textbf{CUB} & \textbf{FLO} & \multicolumn{3}{c}{\textbf{AWA2}} & \multicolumn{3}{c}{\textbf{CUB}} & \multicolumn{3}{c}{\textbf{FLO}}  \\
  \cmidrule(lr){3-3} \cmidrule(lr){4-4} \cmidrule(lr){5-5} \cmidrule(lr){6-8} \cmidrule(lr){9-11} \cmidrule(lr){12-14}
   \textbf{} & & \textbf{T1} & \textbf{T1} & \textbf{T1}& \textbf{u} & \textbf{s} & \textbf{H} & \textbf{u} & \textbf{s} & \textbf{H} & \textbf{u} & \textbf{s} & \textbf{H}  \\
  	\midrule
  	\texttt{GloVe}~\cite{glove} & \texttt{CLSN} & 52.1 & 20.4 & 21.6 & 42.1 & 75.3 & 54.0 & 16.2 & 43.6 & 23.6 & 14.4 & 88.3 & 24.8 \\
  	\texttt{GloVe}~\cite{glove} & \texttt{Wiki} & 61.6 & 29.0 & 25.8 & 49.5 & 78.1 & 60.6 & 23.8 & \textbf{62.6} & 34.5 & 14.7 & 91.0 & 25.3\\
  	\texttt{LongFormer}~\cite{Beltagy2020Longformer} & \texttt{Wiki} & 44.2 & 22.6 & 8.8 & 41.6 & 81.8 & 55.2 & 19.9 & 41.0 & 26.8 & 8.8 & 89.8 & 16.0 \\
  	\texttt{MPNet}~\cite{mpnet} & \texttt{Wiki} & 61.8 & 25.8 & 26.3 & 58.0 & 76.4 & 66.0 & 20.6 & 44.3 & 28.2 & 22.2 & \textbf{96.7} & 36.1 \\
  	\texttt{TF-IDF}~\cite{tfidf} & \texttt{Wiki} & 46.4 & 39.9 & 34.0 & 29.6 & \textbf{87.6} & 44.2 & 29.0 & 52.1 & 37.3 & 28.9 & 94.8 & 44.3 \\
  	\texttt{VGSE}~\cite{vgse} & \texttt{IMG} + \texttt{CLSN} & 69.6 & 37.1 & - & 56.9 & 82.8 & 67.4 & 27.6 & 70.6 & 39.7 & - & - & -\\
  	\midrule
  	\multirow{3}{8em}{\texttt{I2DFormer}~\cite{naeem2022i2dformer}}& \texttt{Wiki} & 
  	{76.4} & {45.4} & {40.0} & 
  	{66.8} & {76.8} & {71.5} & 
  	{35.3} & {57.6} & {43.8} & 
  	{35.8} & {91.9} & {51.5} \\ 
  	& \texttt{3-LLM} (ours) & 
  	{69.7} & {46.0} & {41.9} & 
  	{65.2} & \underline{80.4} & {72.0} & 
  	{36.6} & \underline{59.5} & {45.3} & 
  	{37.4} & \underline{94.2} & {53.5} \\ 
  	& \texttt{3-LLM + Wiki} (ours) & 
  	\underline{77.3} & \underline{47.0} & \underline{43.0} & 
  	\underline{68.6} & {77.4} & \underline{72.7} & 
  	\underline{38.5} & {59.3} & \underline{46.7} & 
  	\underline{40.4} & {80.1} & \underline{53.8} \\  
  	\midrule
    \multirow{3}{8em}{\textbf{\ours}~(ours)} & \texttt{Wiki} & 
  	{73.6} & {42.1} & {41.3} & 
  	{66.6} & \underline{82.9} & {73.8} &
  	{32.4} & \underline{63.1} &  {42.8} & 
  	{34.9} & {\underline{96.1}} & {51.2} \\
  	& \texttt{3-LLM} (ours) & 
  	{76.4} & {47.8} & {44.4} & 
  	{{72.7}} & {81.3} & {{76.8}} &
  	{40.1} & {58.0} & {47.4} & 
  	{{41.1}} & {91.1} & {{56.6}} \\
  	& \texttt{3-LLM + Wiki} (ours) & 
  	\textbf{\underline{79.6}} & \textbf{\underline{51.1}} & \textbf{\underline{46.2}} & 
  	\textbf{\underline{75.7}} & {{79.6}} & \textbf{\underline{77.6}} & 
  	\textbf{\underline{42.5}} & {{59.9}} & \textbf{\underline{49.7}} & 
  	\textbf{\underline{41.6}} & {91.0} & \textbf{\underline{57.1}} \\
  	\bottomrule
  \end{tabular}
}
\caption{\textbf{Comparing our \ours with baseline.} Our \ours significantly improves on the baselines to set a new SOTA for unsupervised class embeddings. We report top-1 accuracy~(\textbf{T1}) on unseen classes for ZSL, and seen/unseen~(\textbf{s/u}) classes and their harmonic mean~(\textbf{H}) for GZSL. We see that the 3-LLM generated views provide complementary information to the wiki articles and significantly improve the performance.
Moreover, we see that \ours is better at consuming multi-view knowledge compared I2DFormer. 
Finally, we see that \ours with LLM supervision alone can outperform I2DFormer with Wiki article indicating that the LLM alone can generate rich class descriptions. Best results within a method are \underline{underlined}. Best results overall are in \textbf{bold}. 
}
\vspace{-10pt}
\label{tab:i2dcomparison}
\end{table*}

\section{Experiments}
We conduct extensive experimentation on three popular ZSL datasets Animal with Attributes 2~(AWA2)~\cite{32_awa}, Caltech-UCSD Birds~(CUB)~\cite{26_wah2011caltech} and Oxford Flowers~(FLO)~\cite{OxfordFlowersDataset} using the evaluation protocol and data splits proposed by Xian et al.~\cite{xian2018zero}. We do not use any human-labeled attributes similar to other works in unsupervised class embeddings. In the following, we discuss implementation details, detailed experiments and 
their conclusions.

\myparagraph{Implementation Details.} We use PaLM540B~\cite{chowdhery2022palm} as the Large Language Model (LLM) for our main experiments prompted with two shots per view at a temperature value of 0.9. 
We use 3 LLM generated views in addition to the wiki articles released by \cite{naeem2022i2dformer} for our main experiments. 
The \ourssummary is implemented as a two-block deep text transformer. The MVSummary transformer is implemented with a similar configuration to \ourssummary. $T$ the number of summary tokens is set as 64 for CUB and 128 for AWA and FLO. We use GloVe~\cite{glove} as the initial token representation similar to \cite{naeem2022i2dformer}. We use the VIT/B16 checkpoint trained on ImageNet1K as the pretrained Image Transformer to be consistent with previous work. The patch projection and MLP in \ourssummary are two layers with ReLU and LayerNorm. For GZSL, we apply calibrated stacking~\cite{chao2016empirical} to calibrate the activations of unseen classes on a held-out set. We use the Adam optimizer with a learning rate of 1$e^{-3}$ and the model converges in $\approx$ 24 hours. $L_{CLS}$ and $L_{Local}$ are combined with weights ablated on the validation set. Detailed training details and examples of LLM-generated views are available in the supplementary. 
Our experimentation framework is implemented in PyTorch and the model can be trained on a single A100 40GB GPU. For the previous SOTA, I2DFormer, we concatenate the text in different views as it is designed for a single view (document) of a class.
Performance of pretrained/ classical embedding baselines like GloVe~\cite{glove}, TF-IDF~\cite{tfidf}, VGSE~\cite{vgse} etc. are taken from \cite{naeem2022i2dformer} with their training setup. We report the top-1 per-class mean accuracy in ZSL. In GZSL, we report the top-1 per-class mean accuracy on seen (s) and unseen (u) classes separately along with their harmonic mean (H).

\subsection{Comparing with State-of-the-Art.}
We compare our results with state-of-the-art in unsupervised class embeddings in Table~\ref{tab:i2dcomparison} and show that LLM-generated multiple views can significantly improve the performance in ZSL. Moreover, we show that \ours significantly outperforms all the previous methods to set a new state-of-the-art in unsupervised class embeddings on all three datasets. Our detailed observations are as follows.

\myparagraph{LLM documents vs Wiki documents.} We observe that Wiki articles from 
\cite{naeem2022i2dformer} and the 3-LLM generated views provide complementary information to consistently improve the ZSL and GZSL performance on all datasets across all metrics. This validates our hypothesis that ZSL models can benefit from multiple perspectives of a class and the LLM is able to generate them without significant annotation effort.
Compared to previous SOTA results of I2DFormer with Wiki, \ours achieves an absolute improvement of 3.2\% on AWA, 5.7\% on CUB and 6.2\% on FLO in ZSL. Similar improvements are seen in the GZSL setting, where we see consistent improvements.
We observe that \ours is better at consuming multi-view knowledge compared to the previous SOTA I2DFormer validating our hypothesis that per-view processing of text allows for extracting richer information from each view. Since our model is specifically developed for multi-view documents, we see that it is on par with I2DFormer across a single view of the Wiki article but achieves significant improvement once multiple views are introduced. 
Finally, we see that \ours with only LLM generated views surpasses I2DFormer with Wiki documents on the three datasets indicating that LLM alone can generate highly discriminative class descriptions for zero-shot image classification. This confirms our hypothesis that LLM with targeted prompting can provide multiple highly discriminative views of a class.

\myparagraph{Learning per-view summary vs text concatenation.} \ours processes each document in the multiple views independently before generating class level $V_{CLS}$ and local multi-view summary tokens $\vi_{\text{mv}}$. This is in contrast to I2DFormer, which concatenates all views into a single text sequence for global and local alignment. We see that the \ours strategy remains superior as the model can first extract highly discriminative facts from each view and later combine them together in the learned $\mathtt{CLS}$ and local summary tokens. When all views are concatenated, the large text sequence can contain repeated information. Moreover, learning local alignment on such a large sequence works less optimally than allowing the model to first extract a set of highly discriminative local summary tokens. We see that \ours consistently outperforms I2DFormer at LLM across 3 and LLM+Wiki supervision across 4 views. These improvements are in addition to the reduced memory cost as \ours requires half the GPU memory compared to I2DFormer across 4 views.  


{
\setlength{\tabcolsep}{4pt}
\renewcommand{\arraystretch}{1.2} 
\begin{table}[t]
\setlength{\aboverulesep}{0pt}\setlength{\belowrulesep}{0pt}
    \centering
    {
    \begin{tabular}{l cccc ccc}
    \toprule
    & \multicolumn{4}{c}{\textbf{Components}} & \textbf{AWA} & \textbf{CUB} & \textbf{FLO}\\
    \cmidrule(lr){6-6} \cmidrule(lr){7-7} \cmidrule(lr){8-8}
    & $L_{CLS}$ & $L_{Local}$ & SVS & MVS &
    \textbf{T1} & \textbf{T1} & \textbf{T1} \\
    \midrule
    a) & \checkmark& &        &         &      73.6     &   45.6 & 38.9\\
    b) & \checkmark& &  \checkmark &  &  {74.1} & 48.5 &  {39.1}  \\
    c) &  & \checkmark & \checkmark &  \checkmark&  57.7 & 32.5 & 24.2  \\
    d) & \checkmark & \checkmark & \checkmark &  & 78.4 & 49.0 & 43.2 \\
    e) & \checkmark & \checkmark & \checkmark & \checkmark & \textbf{79.6} & \textbf{51.1}& \textbf{46.2}  \\
    \bottomrule
    \end{tabular}}
    \vspace{-5pt}
    \caption{\textbf{Ablating over \ours}, we confirm the importance of each component of our model. We observe that $L_{CLS}$ and $L_{Local}$ are complementary to each other. Moreover, SVSummary (SVS) and MVSummary(MVS) reduce the complexity of cross-modal attention while improving performance.}
    \label{tab:ablateloss}
    \vspace{-10pt}
\end{table}
}
\subsection{Ablation over \ours.} 
We ablate over the various components of our model in Table~\ref{tab:ablateloss} using LLM+Wiki views. 
Rows a) and b) only optimize for the global feature between the image and text using $L_{CLS}$. Row a) optimizes for the $V_{CLS}$ generated by the concatenation of all views, while Row b) introduces our SVSummary module. We see that learning a per-view summary while reducing the cost of attention in the text, also offers a performance improvement. 
Row c) only optimizes the fine-grained alignment between image patches and multi-view summary tokens using $L_{Local}$. We see that this alone performs worse than the global head as fine-grained alignment is a hard problem to optimize as also noted in previous works~\cite{naeem2022i2dformer, filip}.
Row d) and e) optimize for both the $L_{CLS}$ and $L_{Local}$. We observe that the two losses are complementary and result in a significant improvement in performance as the model aligns the image and text modality with global as well as local features. Row d) uses a concatenation of per-view summary tokens while Row e) uses our MVSummary module to first learn a set of tokens representing a multi-view summary of the class for learning local alignment. We confirm that MVSummary improves the performance of the model while reducing its memory complexity in cross-modal attention.


\subsection{Ablation on generating views from LLM.}
\label{sec:llmablate}
In this section, we study how to generate good text supervision from an LLM for zero-shot image classification. Unless mentioned, we do not use the Wiki article as a view to only study the impact of the LLM.

\myparagraph{Influence of LLM k-shot prompting on performance.}
We study the impact of k-shot prompting in Table~\ref{tab:numshots} on generating 3 views per class. We observe that even in zero-shot prompting in row a), the views generated by the LLM allow for a very competitive model further validating that LLM can serve as a search engine for generating class supervision in ZSL. 
We observe that with 1-shot prompting in row~b), we see an improvement over zero-shot as the LLM is now aware of what sort of information we require per class. For our 2-shot prompting, we test repeatedly querying the model three times with the same 2-shot example in row~c) vs providing a unique combination of the 4 examples in row~d). We observe that while repeating the same 2-shot example generates competitive views, these provide limited additional information. We see the best performance in row~d) where we prompt the model with unique 2-shot examples. This allows the model to combine two labeling styles to generate a combined perspective that contains more information as evidenced by the improved accuracy numbers. 
We expect a further increase in performance if the LLM has access to more unique k-shot examples. However, this would require labeling 10 k-shot examples which are already 20\% of the classes for AWA. Since we are interested in learning semantic embeddings with minimal supervision, we leave studying this for future works.

\setlength{\tabcolsep}{4pt}
\renewcommand{\arraystretch}{1.2} 
\begin{table}[t]
\setlength{\aboverulesep}{0pt}\setlength{\belowrulesep}{0pt}
\centering
\small
 \resizebox{\linewidth}{!}{%
   \begin{tabular}{l l c c c c c c c c }
  	\toprule
  	& & \multicolumn{2}{c}{\textbf{Zero-Shot Learning}} & \multicolumn{6}{c}{\textbf{Generalized Zero-Shot Learning}} \\
  	\cmidrule(lr){3-4} \cmidrule(lr){5-10}
  & \textbf{Shots}  & \makebox[1.2cm][c]{\textbf{AWA2}} & \makebox[1.2cm][c]{\textbf{FLO}} & \multicolumn{3}{c}{\textbf{AWA2}} &  \multicolumn{3}{c}{\textbf{FLO}}  \\
  \cmidrule(lr){3-3} \cmidrule(lr){4-4} \cmidrule(lr){5-7} \cmidrule(lr){8-10}
   \textbf{} & & \textbf{T1} & \textbf{T1}& \textbf{u} & \textbf{s} & \textbf{H} & \textbf{u} & \textbf{s} & \textbf{H}  \\
  	
  	\hline
  	a) & \texttt{0 shot} & 
  	{73.0} & {40.7} & 
  	{66.6} & {79.1} & {72.3} & 
  	{38.0} & {85.7} & {52.7} \\
  	b) & \texttt{1 shot unique} & 
  	{{74.2}} & {42.1} & 
  	{{68.8}} & {\textbf{82.8}} & {{75.1}} & 
  	{{39.8}} & {89.9} & {{55.2}} \\
  	c) & \texttt{2 shots repeated} & 
  	{73.1} & {43.1} & 
  	{67.8} & {79.9} & {73.4} & 
  	{39.7} & {90.1} & {55.1} \\
  	d) & \texttt{2 shots unique} & 
  	{\textbf{76.4}} & {\textbf{44.4}} & 
  	{\textbf{72.7}} & {81.3} & {\textbf{76.8}} & 
  	{\textbf{41.1}} & \textbf{91.1} & {\textbf{56.6}} \\
\bottomrule
  \end{tabular}
}
\vspace{-5pt}
\caption{\textbf{Ablating over different prompting strategies}, we observe that k-shot prompting works better than 0 shot prompting resulting in richer class descriptions. Moreover, unique k-shot examples serve better at generating multiple views than repeated k-shot examples for each view.
}
\label{tab:numshots}
\end{table}

\setlength{\tabcolsep}{4pt}
\renewcommand{\arraystretch}{1.2} 
\begin{table}[t]
\setlength{\aboverulesep}{0pt}\setlength{\belowrulesep}{0pt}
\centering
\small
 \resizebox{\linewidth}{!}{%
   \begin{tabular}{ l  c c c c c c c c }
  	\toprule
  	& \multicolumn{2}{c}{\textbf{Zero-Shot Learning}} & \multicolumn{6}{c}{\textbf{Generalized Zero-Shot Learning}} \\
   \cmidrule(lr){2-3} \cmidrule(lr){4-9}
  \textbf{Views from LLM}  & \makebox[1.2cm][c]{\textbf{AWA2}} & \makebox[1.2cm][c]{\textbf{FLO}} & \multicolumn{3}{c}{\textbf{AWA2}} &  \multicolumn{3}{c}{\textbf{FLO}}  \\
  \cmidrule(lr){2-2} \cmidrule(lr){3-3} \cmidrule(lr){4-6} \cmidrule(lr){7-9}
   \textbf{} & \textbf{T1} & \textbf{T1}& \textbf{u} & \textbf{s} & \textbf{H} & \textbf{u} & \textbf{s} & \textbf{H}  \\
  	
  	\hline
  	\texttt{1} & 
  	{71.6} & {39.0} & 
  	{67.5} & {75.2} & {71.2} & 
  	{34.6} & {88.0} & {49.6} \\
  	\texttt{2} & 
  	{74.8} & {43.6} & 
  	{70.5} & {80.2} & {75.0} & 
  	{37.7} & {91.0} & {53.3} \\
  	\texttt{3} & 
  	{{76.4}} & {{44.4}} & 
  	{{72.7}} & {81.3} & {{76.8}} & 
  	{{41.1}} & \textbf{91.1} & {{56.6}} \\
  	\texttt{3 + Wiki} & 
  	{\textbf{79.6}} & {\textbf{46.2}} & 
  	\textbf{75.7} & \textbf{79.6} & \textbf{77.6} & 
  	\textbf{41.6} & {91.0} & \textbf{57.1} \\
  	\texttt{4} & 
  	{76.6} & {44.5} & 
  	{72.9} & {81.2} & {76.8} & 
  	{40.5} & {89.6} & {55.8} \\
  \bottomrule
  \end{tabular}
}
\vspace{-5pt}
\caption{\textbf{Ablating over number of views}, we observe that each view provides a useful source of information for \ours and improves the model performance.
}
\vspace{-10pt}
\label{tab:numviews}
\end{table} 
\myparagraph{Impact of Multiple views on performance.}
We study the impact of introducing multiple views as supervision for \ours in Table~\ref{tab:numviews}. We observe that increased views generated with each k-shot example consistently improve the performance. The best performance is achieved by the three views and LLM for a total of 4 perspectives about each class available to the model. This further validates our hypothesis that increased views representing different annotator biases can improve the zero-shot performance. The introduction of LLM as an annotator enables it without requiring actual human annotators as the LLM has stored the knowledge available online and can use it to mimic the annotators available in the k-shot example. Comparing the performance of 2 views in Table~\ref{tab:numviews} with using 3 views across the repeated 2-shot example in row c) of Table~\ref{tab:numshots}, we observe that the model generally benefits more from increased perspectives used in the k-shot example of each view than increased views without new perspective in the k-shot example. Moreover, we observe that the fourth view from LLM which repeats the k-shot example performs worse than using the Wiki article as the fourth view. 
This hints that the knowledge retrieval ability of an LLM is impacted by the information available in its k-shot examples. 
We expect there to be a potential further increase in performance if the LLM has access to more unique k-shot examples but this again comes at increased labeling cost.

\begin{figure}[h]
     \centering
     \begin{subfigure}[h]{0.23\textwidth}
         \centering
         \includegraphics[width=\textwidth]{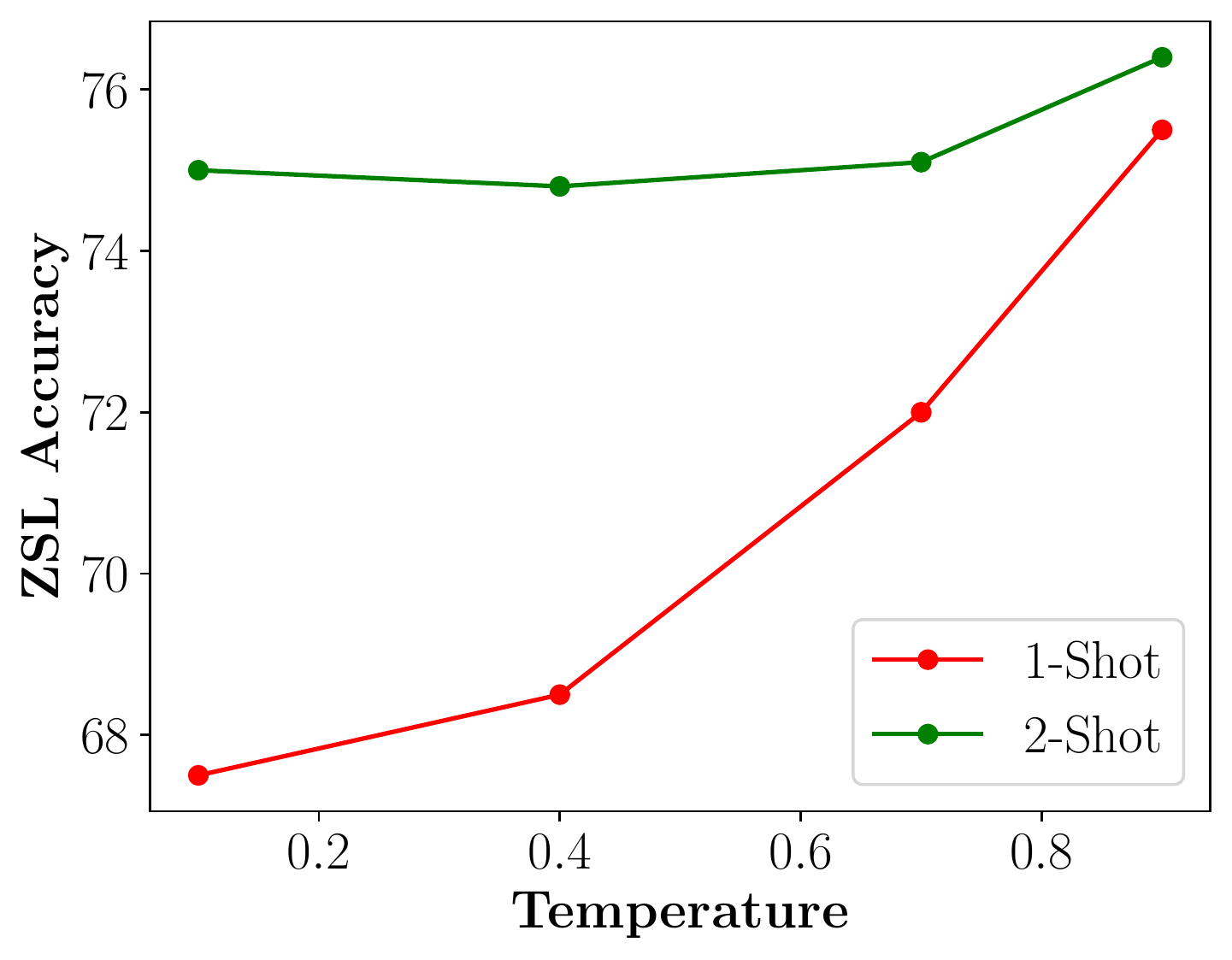}
         \caption{AWA}
     \end{subfigure}
     \begin{subfigure}[h]{0.23\textwidth}
         \centering
         \includegraphics[width=\textwidth]{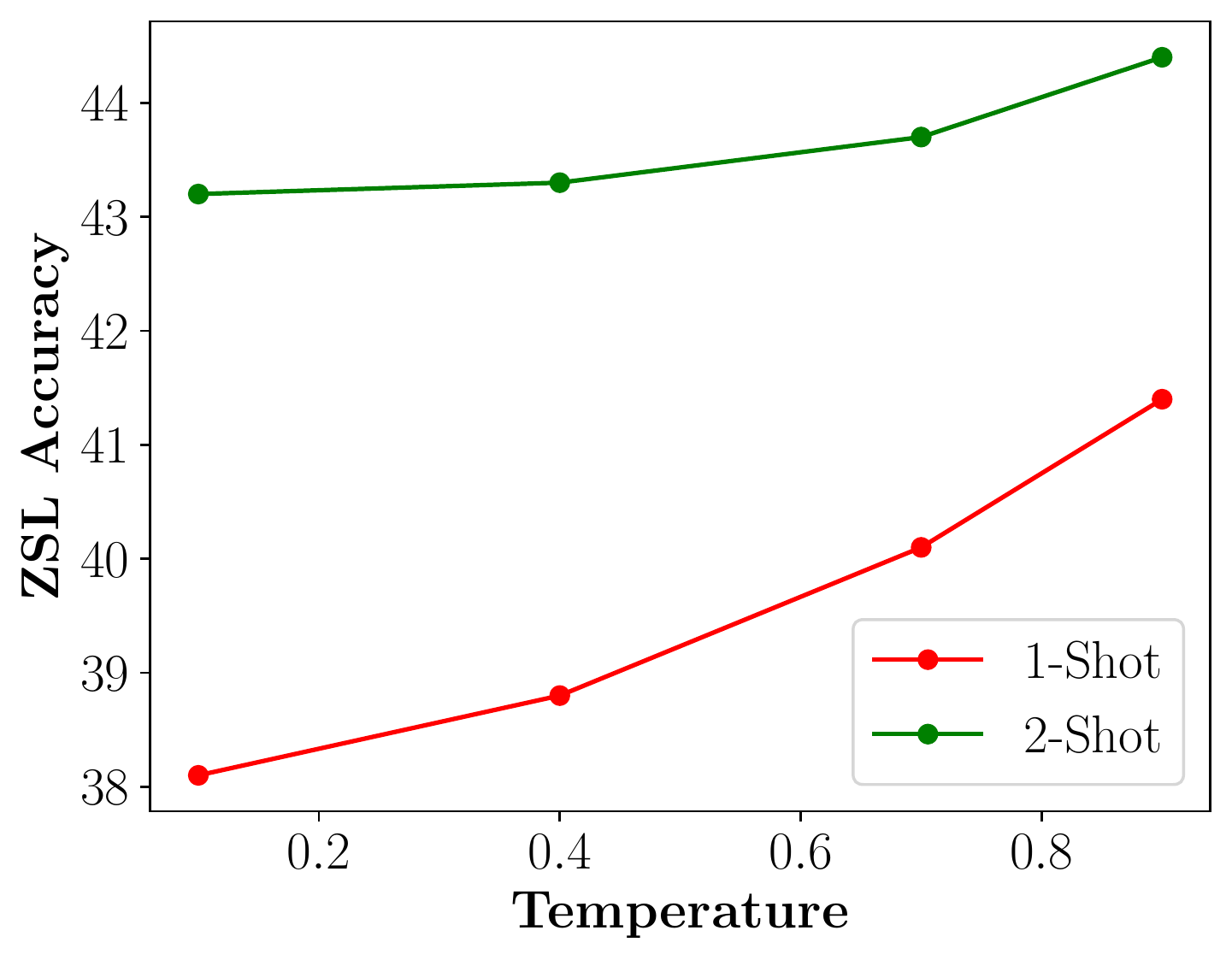}
         \caption{FLO}
     \end{subfigure}
    \vspace{-5pt}
    \caption{\textbf{Ablation over Temperature values.} We observe a performance increase with the temperature value of the LLM. However, at 2-shots, the model is fairly robust to changing temperature and achieves impressive performance at all temperature values.}
    \label{fig:llmtemperature}
    \vspace{-5pt}
\end{figure}


\myparagraph{Impact of temperature value of LLM on performance.}
We study the impact of the temperature value of the LLM in Figure~\ref{fig:llmtemperature}. The temperature value of the LLM controls the distribution it explores. A high value means the language model will sample from further away from the mean and introduce more stochasticity. We observe from Figure~\ref{fig:llmtemperature} that the performance of \ours improves with the temperature value as the LLM can represent more diverse text views of the class with its effect most profound in one-shot prompting. However, this temperature value can end up being yet another hyperparameter to ablate over. We notice that in two-shot prompting, this temperature value has a smaller impact on the performance of \ours as the 2 examples provide better conditioning for each view to constrict the model in what kind of information we require for each class. We see a smaller change in performance across different temperature values with the best performance achieved at a temperature of 0.9 on both datasets. 


\myparagraph{Impact of size and family of LLM on performance.} We study the impact of different LLM in Table~\ref{tab:llmtype} and observe that the largest model PaLM540B provides the best auxiliary information leading to the best performance. Moreover, we observe that the smaller 60B version of PaLM achieves very promising performance indicating that while increased parameters do bring more performance, the ``smaller" LLM can still achieve impressive results. As we compare the results of PaLM with GPT3, we observe that the larger PaLM540B model outperforms GPT3 for our 2-shot setup as also noted in the original PaLM manuscript~\cite{chowdhery2022palm}. 

\setlength{\tabcolsep}{4pt}
\renewcommand{\arraystretch}{1.2} 
\begin{table}[t]
\setlength{\aboverulesep}{0pt}\setlength{\belowrulesep}{0pt}
\centering
\small
 \resizebox{\linewidth}{!}{%
   \begin{tabular}{ l  c c c c c c c c }
  	\toprule
  	& \multicolumn{2}{c}{\textbf{Zero-Shot Learning}} & \multicolumn{6}{c}{\textbf{Generalized Zero-Shot Learning}} \\
   \cmidrule(lr){2-3} \cmidrule(lr){4-9}
  \textbf{LLM}  & \makebox[1.2cm][c]{\textbf{AWA2}} & \makebox[1.2cm][c]{\textbf{FLO}} & \multicolumn{3}{c}{\textbf{AWA2}} &  \multicolumn{3}{c}{\textbf{FLO}}  \\
  \cmidrule(lr){2-2} \cmidrule(lr){3-3} \cmidrule(lr){4-6} \cmidrule(lr){7-9}
   \textbf{} & \textbf{T1} & \textbf{T1}& \textbf{u} & \textbf{s} & \textbf{H} & \textbf{u} & \textbf{s} & \textbf{H}  \\
  	
  	\hline
  	\texttt{PaLM 62B} & 
  	{74.0} & {38.6} & 
  	{66.1} & \textbf{82.3} & {73.3} & 
  	{37.1} & {70.3} & {48.6} \\
  	\texttt{GPT3 175B} & 
  	{74.2} & {44.2} & 
  	{68.8} & {81.0} & {74.2} & 
  	{40.4} & {83.4} & {54.5} \\
  	\texttt{PaLM 540B} & 
  	{\textbf{76.4}} & {\textbf{44.4}} & 
  	{\textbf{72.7}} & {81.3} & {\textbf{76.8}} & 
  	{\textbf{41.1}} & \textbf{91.1} & {\textbf{56.6}} \\
   \bottomrule
  \end{tabular}
}
\vspace{-5pt}
\caption{\textbf{Ablating over different LLM}, we observe that our prompting strategy can be used with different LLMs for generating powerful supervision for zero-shot image classification.
}
\vspace{-12pt}
\label{tab:llmtype}
\end{table}



\section{Conclusion}
\label{sec:conclusion}
We propose a novel perspective of using a Large Language Model as an oracle to reveal multiple views (text descriptions) of a class. Since an LLM is trained on webscale data it only requires a few k-shot examples to generate multiple high-quality text descriptions. 
We show that these LLM-generated views provide complementary information to Wiki documents for learning a zero-shot image classification model.
We propose \ours, a novel transformer-based model, that incorporates our SVSummary module to learn a per-view summary representing discriminative information about a class available in each view. These summaries are used by our MVSummary module to learn class-level multi-view summaries. The multi-view summaries are aligned with the global and local image information to learn a highly discriminative zero-shot image classification model. Our summary modules allow a reduction in the memory requirement of utilizing text in zero-shot image classification models. Moreover, \ours brings significant performance improvements to set a new state of the art in unsupervised semantic embeddings.

\pagebreak
\myparagraph{Limitations of LLM.}
We treat the output of the LLM as factually correct in this work using the accuracy numbers as a proxy. LLM as an annotator to generate auxiliary class information can open research in zero-shot learning in new domains. However, these models come with their own set of biases from their pretraining data. These biases should be carefully studied before introducing LLM-generated text in domains where it can have severe consequences.

{\small
\bibliographystyle{ieee_fullname}
\bibliography{egbib}
}

\end{document}